\documentclass{article}

\usepackage[preprint]{neurips_2025}

\usepackage[utf8]{inputenc} 
\usepackage[T1]{fontenc}    
\usepackage{hyperref}       
\usepackage{url}            
\usepackage{booktabs}       
\usepackage{amsfonts}       
\usepackage{nicefrac}       
\usepackage{microtype}      
\usepackage{xcolor}         
\usepackage{makecell}
\usepackage{microtype}
\usepackage{hyperref}
\usepackage{url}
\usepackage{verbatim}
\usepackage{listings}
\usepackage{booktabs}
\usepackage{graphicx}
\usepackage{lineno}
\usepackage{subcaption}
 \usepackage{multirow} 
\usepackage{amsmath} 
\usepackage[most]{tcolorbox}
\usepackage{fontawesome5} 
\usepackage{xcolor}
\usepackage{amssymb}

\definecolor{promptblue}{RGB}{41, 98, 255}
\definecolor{lightblue}{RGB}{240, 247, 255}
\definecolor{codegray}{RGB}{245, 245, 245}
\definecolor{darkblue}{rgb}{0, 0, 0.5}
\title{Thinking Fast and Right: Balancing Accuracy and Reasoning Length with Adaptive Rewards}

\author{Jinyan Su,~
 Claire Cardie\\
   Cornell University \\
  \texttt{\{js3673, ctc9\}@cornell.edu}
  }

\begin{document}

\maketitle

\begin{abstract}
Large language models (LLMs) have demonstrated strong reasoning abilities in math mathematical reasoning, often enhanced through reinforcement learning (RL). However, RL-trained models frequently produce unnecessarily long reasoning traces—even for simple queries—leading to increased inference costs and latency. While recent approaches attempt to control verbosity by adding length penalties to the reward function, these methods rely on fixed penalty terms that are hard to tune and cannot adapt as the model’s reasoning capability evolves, limiting their effectiveness.
In this work, we propose an adaptive reward-shaping method that enables LLMs to "think fast and right"—producing concise outputs without sacrificing correctness. Our method dynamically adjusts the reward trade-off between accuracy and response length based on model performance: when accuracy is high, the length penalty increases to encourage faster length reduction; when accuracy drops, the penalty is relaxed to preserve correctness. This adaptive reward accelerates early-stage length reduction while avoiding over-compression in later stages. Experiments across multiple datasets show that our approach consistently and dramatically reduces reasoning length while largely maintaining accuracy, offering a new direction for cost-efficient adaptive reasoning in large-scale language models. Our code can be found in \url{https://github.com/JinyanSu1/A-DLP}.
\end{abstract}

\section{Introduction}
Recent advances in large language models (LLMs) have demonstrated impressive reasoning capabilities across domains such as mathematics \citep{cobbe2021training, hendrycks2021measuring} and programming \citep{Codeforces}, often enhanced through reinforcement learning (RL) \citep{guo2025deepseek, openai}. While the long-form reasoning behavior encouraged by RL has substantially improved model performance across benchmarks, it also incurs significant computational overhead and latency.
Moreover, LLMs frequently “overthink” even simple questions \citep{chen2024not, su2025between, sui2025stop}, producing unnecessarily verbose output with thousands of tokens even for questions such as \textit{"what is 2 plus 3?"}. 

To mitigate this, many research efforts have explored ways to encourage more efficient reasoning \citep{sui2025stop, wang2025harnessing, feng2025efficient}. A particularly practical approach is to incorporate length penalties directly into the reward function during RL training to penalize excessively long outputs. Since many reasoning-capable LLMs—such as DeepSeek-R1 \citep{guo2025deepseek}, OpenAI's o1 \citep{openai}, and QwQ-32B-Preview \citep{QwenTeam}—already include RL in their training pipelines, modifying the reward function becomes a lightweight and low-overhead intervention. It integrates seamlessly into existing workflows without requiring architectural changes or additional training infrastructure, making it more practical and scalable than alternative approaches that operate outside the RL framework.

Several recent works have investigated reward modifications to reduce reasoning length. For example, L1 \citep{aggarwal2025l1} modified the reward to penalize responses that exceed user provided token budget; O1-Pruner \citep{luo2025o1} formulates a constrained optimization problem where accuracy is enforced as a hard constraint while minimizing response length; and \cite{yi2025shorterbetter} rewards outputs that are close in length to the shortest correct response. Despite differences in formulation, these methods all depend on a \textbf{fixed-length penalty} parameter to manage the trade-off between accuracy and generation efficiency. However, it is difficult to set a universally effective value: if the penalty is too large, the model may collapse into overly short outputs that sacrifice accuracy; if it is too small, the model may retain unnecessarily long outputs or take too long to converge to a reasonable response length.

To address this limitation, we propose a simple and intuitive approach that adaptively adjusts the reward trade-off between response length and accuracy  based on model performance. Specifically, during training, we continuously monitor the model's accuracy relative to a reference accuracy threshold and dynamically adjust the length penalty: when accuracy exceeds the threshold, we increase the length penalty to encourage faster reduction in response length; when accuracy falls below the threshold, we relax the penalty to preserve correctness.
Crucially, our method allows the penalty to evolve and adapt with the model’s capabilities. In the early stages, when the model outputs are excessively verbose, a stronger penalty should be used to eliminate redundancy and accelerate length compression. As training progresses and the model begins to produce more concise outputs, continuing to apply a strong penalty can lead to overly aggressive attempts to shorten responses as well asdegraded performance. By allowing the penalty to decrease when needed, the model can shift its focus toward preserving correctness rather than overly compressing the response.

To summarize, our contributions are:
\begin{itemize}
\item We propose an Adaptive Direct Length Penalty (A-DLP) reward that evolves throughout training to reduce response length while maintaining acceptable accuracy. By adaptively
 shaping the reward signal based on model performance, our method enables fast compression in the early stages and prevents over-compression in later stages.
\item Through extensive experiments, we demonstrate the effectiveness and advantages of our approach. Our results consistently show improved generation efficiency without compromising performance, and our analyses offer new insights on how adaptive reward shaping can better balance the trade-off between reasoning length and accuracy. 
\end{itemize}
Our adaptive framework opens a promising direction for future research in building more cost-effective and efficient reasoning language models.

\section{Related Work}
\paragraph{LLM Reasoning} LLMs have
demonstrated impressive performance on a wide range of reasoning tasks \citep{roziere2023code, rein2024gpqa, cobbe2021training, hendrycks2021measuring} particularly when using Chain-of-Thought (CoT) prompting \citep{wei2022chain}. To further improve reasoning performance, a number of test-time techniques have been proposed that increase computational cost in exchange for better accuracy. For example, self-consistency \citep{wang2022self} samples multiple CoT traces and selects the most frequent final answer, while Best-of-N sampling \citep{gao2023scaling} selects the best response from several samples based on a scoring function. Process reward models \citep{lightman2023let} evaluate CoT progress step-by-step, while self-correction methods \citep{kumar2024training} refine initially generated responses through iterative feedback. Although these test-time methods improve performance, modern large-scale reasoning models such as DeepSeek-R1 \citep{guo2025deepseek}, OpenAI o1 \citep{openai}, and QwQ-32B-Preview \citep{QwenTeam} are trained with reinforcement learning (RL), leading to the emergence of structured behaviors such as verification, backtracking, and branching in long CoT traces \citep{wu2025unlocking}.

\paragraph{Long-to-Short Reasoning.}
While deliberate reasoning is useful for complex tasks, it often results in inefficient or unnecessarily verbose outputs. This phenomenon, referred to as the "overthinking problem" \cite{chen2024not, su2025between, yang2025towards}  arises when increased reasoning length fails to yield better answers. To address this, a growing body of work explores methods for shortening CoT reasoning \citep{sui2025stop, wang2025harnessing, feng2025efficient}. For example, C3oT \citep{kang2025c3ot} trains a compressor to convert verbose CoT traces. Others focus on prompt-based methods: \citep{nayab2024concise, han2024token} attempt to explicitly constrain token length through prompt engineering. TokenSkip \citep{xia2025tokenskip} enables LLMs to selectively skip over less informative tokens. Meanwhile, \cite{munkhbat2025self} uses Best-of-N sampling to construct concise data for fine-tuning. For comprehensive overviews, we refer readers to recent surveys on efficient reasoning \citep{sui2025stop, wang2025harnessing, feng2025efficient}.
\paragraph{Reasoning Length Reduction via RL.} Several recent works leverage RL to directly reduce reasoning length by modifying the reward function. \cite{hou2025thinkprune}  uses the correctness of clipped responses as the reward. L1 \citep{aggarwal2025l1} adds a penalty term to enforce generation length budgets specified in the prompt. O1-Pruner \citep{luo2025o1} and \citep{arora2025training} penalize the reward with normalized length to ensure proportional compression. \cite{yi2025shorterbetter} propose a reward that penalizes deviations from the shortest correct responses among the samples. However, all of these methods rely on a fixed penalty parameter to control the trade-off between accuracy and length, which must be manually tuned and does not adapt to the model’s training dynamics.

\section{Method}

When fine-tuning large language models (LLMs) using reinforcement learning (RL), the objective is to optimize the policy $\pi_\theta$ such that the generated response $y \sim \pi_\theta(\cdot \mid x)$ is correct. Let $y^*$ denote the reference answer associated with input $x$. A basic reward function focused on correctness can be written as:
\begin{equation}
R_{\text{correct}}(x, y) = \mathbb{I}\{y = y^*\}
\end{equation}
where $\mathbb{I}$ is the indicator function. While this encourages accurate generation, it does not consider the efficiency or length of the response—an important factor in practical deployment.

\subsection{Static Direct Length Penalty (S-DLP)}
To encourage concise reasoning, it is natural to introduce a penalty on the reward for response length. Let $\text{len}(y)$ denote the number of tokens in the generated output. A straightforward formulation, which we refer to as the Static Direct Length Penalty (S-DLP), is defined by the following reward:
\begin{equation}\label{eq: 1}
R_\lambda(x, y) = \mathbb{I}\{y = y^*\} - \lambda \cdot \text{len}(y)
\end{equation}
where $\lambda \geq 0$ is a hyperparameter that controls the trade-off between correctness and brevity. A large $\lambda$ encourages shorter responses and accelerates length reduction, but may cause the model to collapse into overly short, incorrect outputs. In contrast, a small $\lambda$ prioritizes correctness, but may result in slow or ineffective reduction in generation length. 
Selecting an appropriate value for $\lambda$ is challenging, as it depends both on the degree of redundancy in the model's current outputs and on the desired trade-off between accuracy and efficiency.
\subsection{Adaptive Direct Length Penalty (A-DLP)}
To avoid static tuning, we propose an adaptive variant of the above reward, referred to as the Adaptive Direct Length Penalty (A-DLP).
In this formulation, the penalty coefficient $\lambda_t$ is dynamically updated throughout training based on the model's observed performance. In particular, during training, we continuously compare the model's accuracy against a reference accuracy: when the model maintains high correctness, we increase the penalty to encourage faster reduction in generation length; when performance drops, we ease the penalty to prioritize preserving accuracy and avoid overly aggressive compression.

Let $\text{acc}_t$ be the observed accuracy at training step $t$, and let $\text{acc}_{\text{ref}}$ be the reference accuracy. We update the penalty coefficient as:
\begin{equation} \label{eq: 2}
\lambda_{t+1} = \max\left(0, \lambda_t + \eta \cdot (\text{acc}_t - \text{acc}_{\text{ref}})\right)
\end{equation}
where $\eta > 0$ is learning rate that controls the sensitivity of the penalty to performance changes. While it is ideal to use a reference accuracy computed from a separate model or evaluation run, doing so would incur additional computational cost—something we aim to avoid.  To eliminate this overhead, we instead set $\text{acc}_{\text{ref}} = \mathbb{E}_{x\sim \mathcal{D}, y\sim \pi_{\text{ref}}(y|x)}[\mathbb{I}\{y=y^{*}\}]$ to the expected accuracy over the data distribution under a reference model. This global constant serves as a fixed performance baseline. In practice, it can be estimated by the accuracy of the reference model before any length reduction is applied. This simplification removes the need for additional model inference during training, while still providing a meaningful anchor for adaptive reward adjustment. The resulting adaptive reward function used in A-DLP then becomes:
\begin{equation} \label{eq: 3}
R_{\lambda_t}(x, y) = \mathbb{I}\{y = y^*\} - \lambda_t \cdot \text{len}(y)
\end{equation}

This formulation allows the reward signal to  evolve adaptively throughout training. In the early stages, when the model's responses are highly redundant, reducing generation length has little to no impact on accuracy and $\lambda_t$ is large enough to ensure rapid compression of unnecessary tokens. As the model continues to shorten its outputs, further reductions may begin to affect correctness. As a result, $\lambda_t$ naturally decreases, relaxing the pressure to shorten responses in order to preserve accuracy. Eventually, the generation length stabilizes at a point where further compression would lead to a measurable drop in accuracy, and $\lambda_t$ converges toward zero. This dynamic adaptation ensures that length reduction proceeds aggressively when safe, and conservatively when needed, without manual tuning.

\section{Experimental Setup}
\paragraph{Models and Datasets} We train our models on the DeepScaleR-Preview-Dataset \citep{deepscaler2025}, a mathematics dataset containing 40K question-answer pairs sourced from AIME, AMC, Omni-Math \citep{gao2024omni}, and STILL \citep{min2024imitate}. We use AIME2024 as the validation set and evaluate on five datasets: AIME2025, MATH \citep{hendrycks2021measuring}, AMC, Olympiad-Bench \citep{he2024olympiadbench}, and Minerva. Our base model is DeepScaleR-1.5B-Preview \citep{deepscaler2025}, a reasoning-oriented language model fine-tuned from DeepSeekR1-Distill-Qwen-1.5B \citep{guo2025deepseek}. For both training and evaluation, we restrict the maximum context length to 8192 tokens.
\paragraph{Baselines} 
We compare our method against the following baselines, selected for their adaptability and availability of released models:
\begin{itemize}
\item \textbf{Base Model}: the original DeepScaleR-1.5B-Preview model, without any length penalization.
\item \textbf{L1-Exact} \citep{aggarwal2025l1}: Fine-tunes DeepScaleR-1.5B-Preview with RL using a reward function that encourages the generation length to match a user-specified target.
\item \textbf{L1-Max} \citep{aggarwal2025l1}: Fine-tunes DeepScaleR-1.5B-Preview using RL with a reward function that penalizes outputs exceeding a specified length budget.
\item \textbf{Static Direct Length Penalty (S-DLP)}: Applies a static penalty on generation length during RL, using Equation~\ref{eq: 1} as the reward function.
\item \textbf{Adaptive Direct Length Penalty (A-DLP)}: Adaptively updates the length penalty parameter $\lambda_t$, and consequently the reward function, based on the model’s current performance, following Equations~\ref{eq: 2} and~\ref{eq: 3}.
\end{itemize}
\paragraph{Implementation Details} We use the standard RL training pipeline for DeepScaleR-1.5B-Preview, applying Group Relative Policy Optimization (GRPO) \citep{shao2024deepseekmath} for policy updates. All experiments are conducted on two A100 GPUs (80GB each). We use a batch size of 64 and constrain the maximum prompt length to 1024 tokens. The actor learning rate is set to $1\times10^{-6}$, with a rollout count of 4. For length penalty parameters, we initialize the penalty coefficient to be $\lambda_0 = 1e-3$, and use a learning rate $\eta = 1e-3$ for updating $\lambda_t$. Training proceeds until the generation length stabilizes. We use the final checkpoint (at step 420) for evaluation. We set the reference accuracy to be 0.62, which is estimated from the training accuracy of the first batch on the base model. We use both the accuracy and average token count as our evaluation metrics, computed over 16 sampled completions per question using temperature 0.6 and top-$p$ sampling with $p=0.95$.
\section{Results and Analysis}
\subsection{Main Results}
\begin{figure}[h]
    \centering
    \includegraphics[width=1\linewidth]{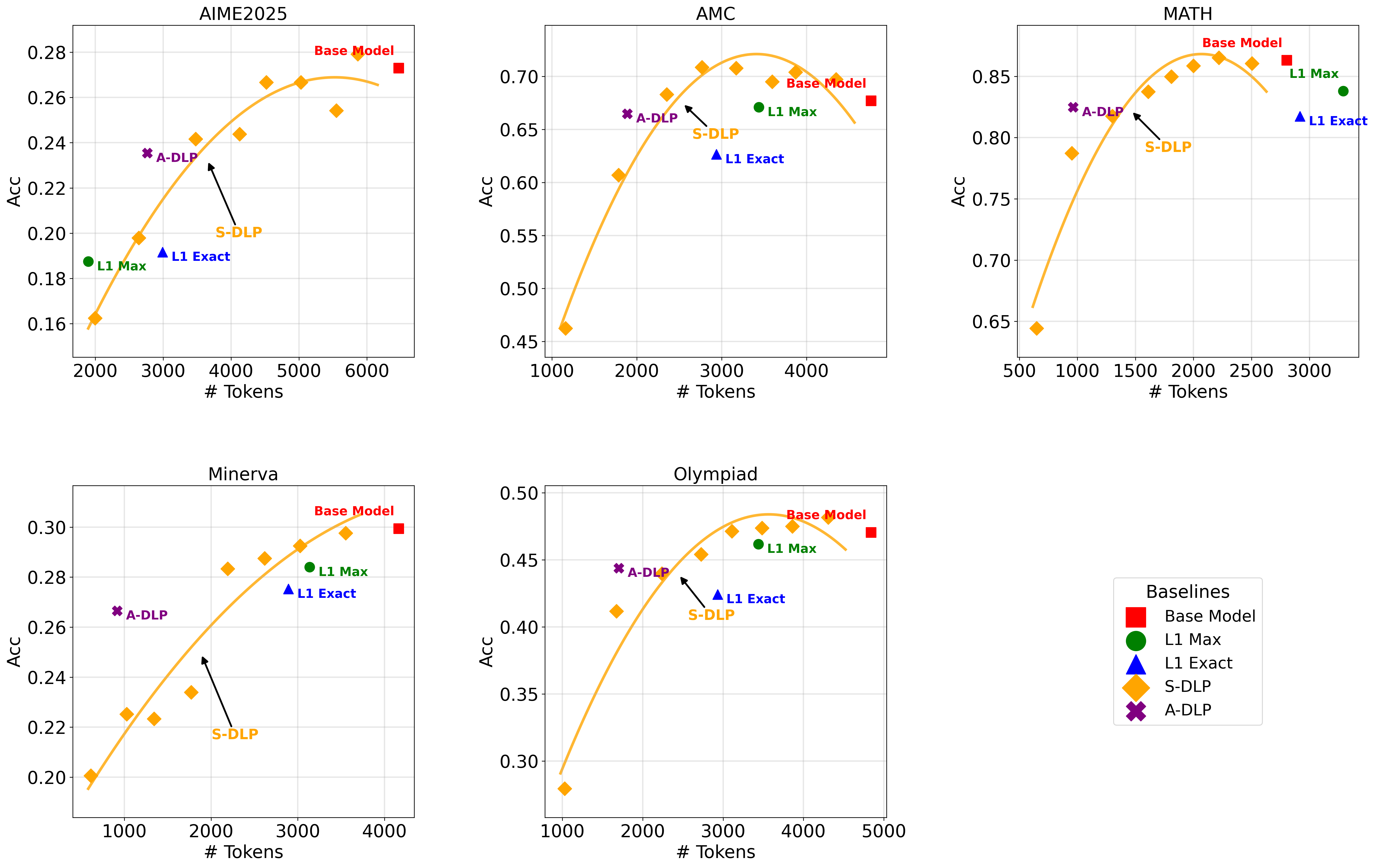}
    \caption{\textbf{Performance comparison of A-DLP with baseline methods.} For the S-DLP, we plot  checkpoints sampled every 20 training steps and fit a curve through them. Since both accuracy and generation length change monotonically during training under S-DLP, this trajectory captures the full accuracy–length trade-off. A-DLP consistently achieves better trade-offs, lying above and to the left of the S-DLP curve. }
    \label{fig: main}
\end{figure}

In Figure \ref{fig: main}, we compare our A-DLP reward against several baselines. For the S-DLP reward baseline, due to the monotonic reduction in generation length over training steps, to better capture the trade-off between accuracy and token length, we plot the full training trajectory by sampling checkpoints every 20 steps and fitting a curve through these points. Across nearly all datasets, A-DLP consistently lies above and to the left of the S-DLP curve, indicating improved length-efficiency, i.e., higher accuracy with fewer tokens. Meanwhile, compared to the original base model, A-DLP reduces token length by more than 50\% while sacrificing less than 0.04 in accuracy. The \textbf{L1-Max} and \textbf{L1-Exact} baselines lie mostly below the S-DLP curve, as their primary objective is to enforce length control around a fixed budget rather than to minimize length for efficiency. Consequently, they tend to prioritize meeting the target length constraint over aggressively reducing generation length when possible. 

\begin{figure}[h]
    \centering
    \includegraphics[width=1\linewidth]{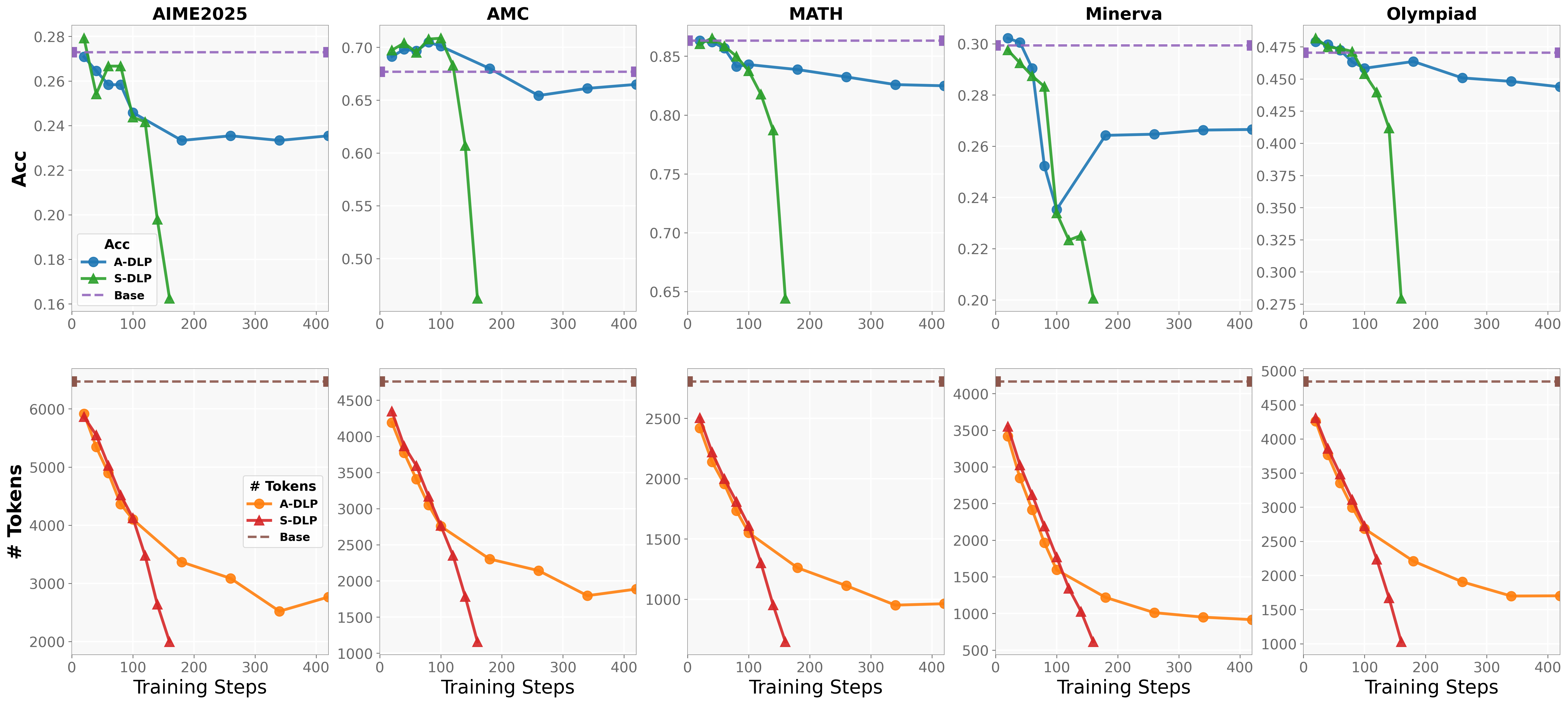}
    \caption{\textbf{Accuracy and average token length across training steps for A-DLP and S-DLP.} The dotted line is the accuracy and token length for the base model before length reduction. For S-DLP, performance remains stable during the early training phase, but both accuracy and token length drop sharply around step 100, indicating model collapse due to excessive length penalization. In contrast, A-DLP exhibits stable convergence, with both metrics gradually stabilizing—demonstrating its ability to adaptively balance correctness and brevity throughout training.}
    \label{fig: training steps}
\end{figure}

\begin{figure}[h]
    \centering
    \includegraphics[width=1\linewidth]{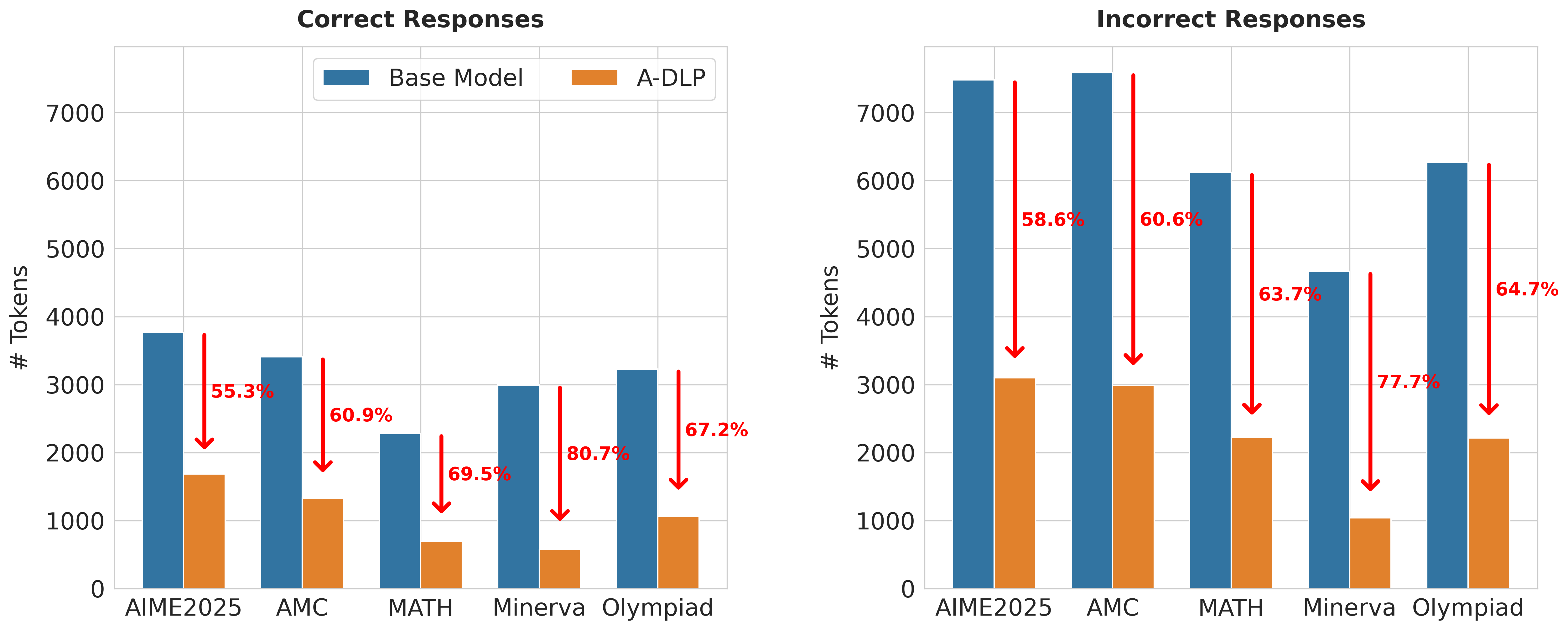}
    \caption{\textbf{Token length of correct and incorrect responses before and after applying A-DLP.} The reduction rates for both categories consistently exceed 55\%.}
    \label{fig: token length comparison}
\end{figure}

Figure~\ref{fig: training steps} provides a detailed comparison of S-DLP and A-DLP on accuracy and token length throughout training. Initially, both S-DLP and A-DLP exhibit similar rates of length reduction while maintaining accuracy close to that of the base model. As training progresses, both methods experience a slight but acceptable drop in accuracy, accompanied by continued reductions in generation length. However, beyond approximately 100 training steps, the behavior of the two methods diverges. Under S-DLP, further training leads to a sharp decline in both accuracy and token length, indicating model collapse due to excessive penalization. In contrast, A-DLP demonstrates stable convergence: both accuracy and token length gradually level off, even as training continues. Thus, A-DLP demonstrates greater robustness with respect to training termination, as it naturally stabilizes without requiring manual intervention to prevent collapse.
In Figure \ref{fig: token length comparison}, we present the token length reduction rates for both correct and incorrect responses. Although incorrect responses are initially much longer than correct ones, our algorithm effectively reduces the token lengths for both categories. Notably, the reduction appears roughly proportional to their original lengths, such that correct responses remain approximately half the length of incorrect ones before and after reduction. Furthermore, the reduction rates for both correct and incorrect responses consistently exceed 55\% across all datasets, demonstrating the effectiveness of our approach.
\subsection{Behavior over Training}
\begin{figure}[h]
    \centering
    \includegraphics[width=1\linewidth]{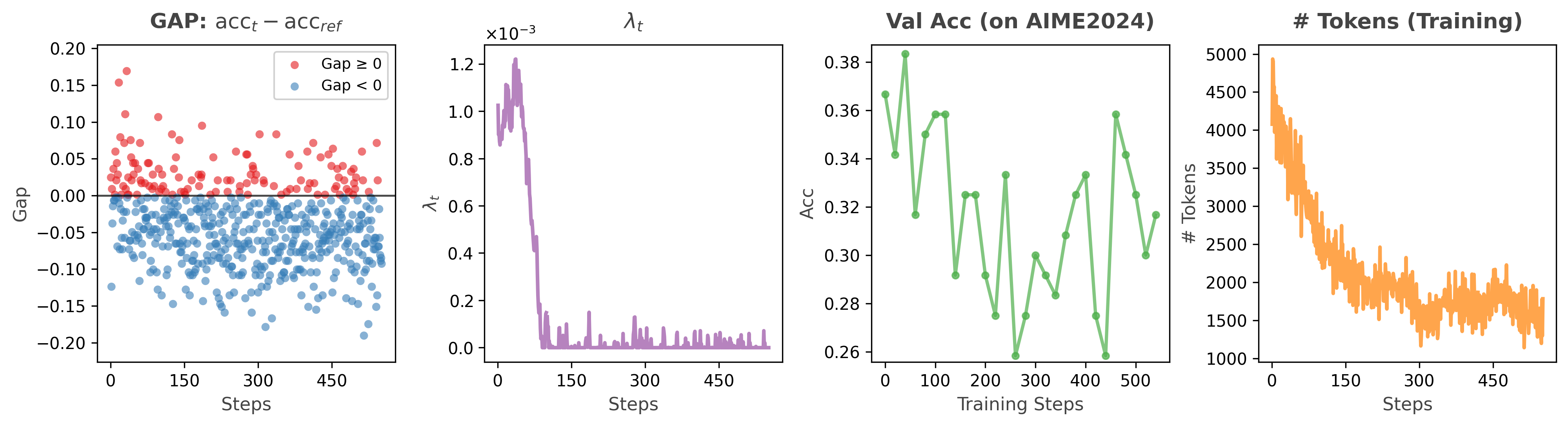}
    \caption{Training dynamics of A-DLP showing the accuracy gap between the current model and the reference threshold ($\text{acc}_t - \text{acc}_{\text{ref}}$), the length penalty coefficient $\lambda_t$, validation accuracy on AIME2024, and the average response length (number of tokens) on the training data.}
    \label{fig: training dynamics}
\end{figure}

Figure~\ref{fig: training dynamics} illustrates the training dynamics of A-DLP, focusing on four key quantities: the accuracy gap between the current model and the reference threshold, i.e., $\text{acc}_t - \text{acc}_{\text{ref}}$; the length penalty coefficient $\lambda_t$; the validation accuracy on AIME2024; and the average response length (i.e., the number of generated tokens) on the training data.
In the early stages of training (before approximately step 50), the model frequently exceeds the reference accuracy, leading to a cumulative increase in the penalty coefficient $\lambda_t$. As training progresses, the model's performance begins to fluctuate around the reference threshold and more often falls below it,  causing $\lambda_t$ to gradually decrease. By around step 100, $\lambda_t$ reaches zero. Beyond this point, it remains near zero, exhibiting only minor oscillations due to noise in the gap estimation.

Correspondingly, the average response length steadily decreases and converges to the range of 1500–2000 tokens—representing more than a 50\% reduction from the initial average of 5000 tokens. Meanwhile, the validation accuracy initially drops slightly as the model aggressively shortens its outputs, but stabilizes around 0.3 once training converges. These patterns demonstrate A-DLP’s capacity to dynamically modulate its penalty signal in response to performance feedback, ultimately achieving a balance between brevity and correctness.

\subsection{Parameter Setting}
In this section, we investigate the effect of tuning core parameters—namely the learning rate $\eta$ and the reference accuracy $\text{acc}_{\text{ref}}$—on model performance. These experiments serve to guide practical choices for parameter selection.

\begin{figure}[h]
    \centering
    \includegraphics[width=1\linewidth]{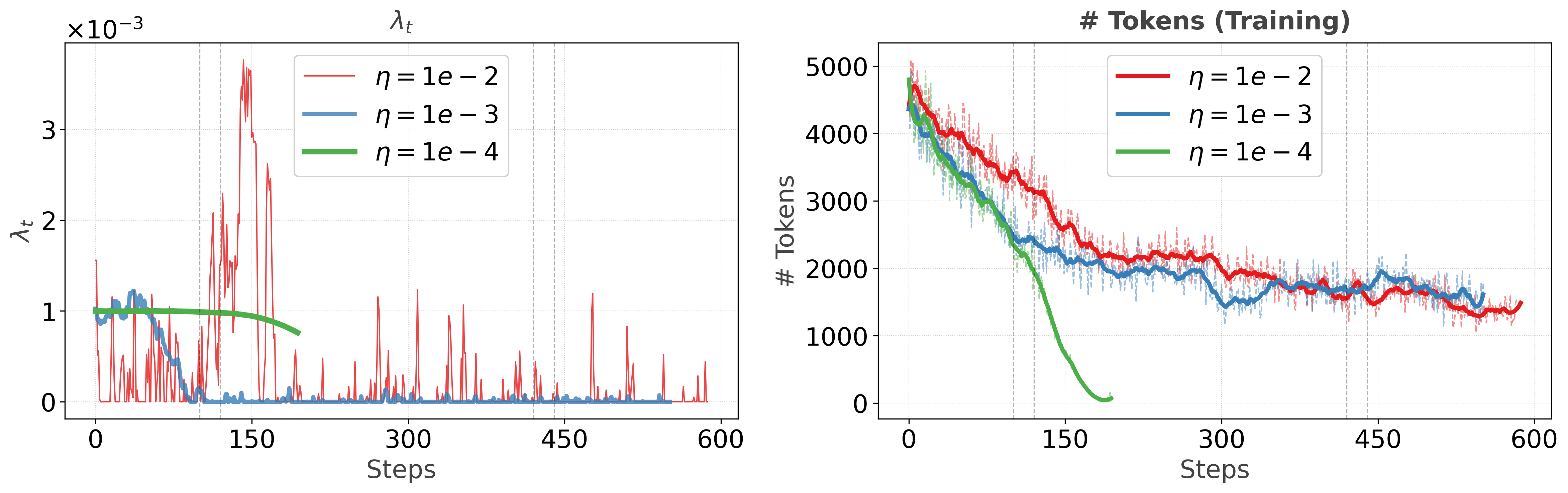}
       \caption{\textbf{Training dynamics of $\lambda_t$ and response length under different learning rates ($\eta \in \{10^{-2}, 10^{-3}, 10^{-4}\}$).} A larger learning rate causes $\lambda_t$ to fluctuate more sharply due to sensitivity to noisy accuracy estimates, resulting in slower length reduction but eventually converges with sufficient training. In contrast, a smaller learning rate leads to smoother updates and faster token reduction in the early training stage, but risks model collapse in later stages, as $\lambda_t$ fails to decrease quickly enough in response to dropping accuracy—causing continued over-penalization and excessive length reduction.}
    \label{fig: different learning rate}
\end{figure}

\paragraph{Effect of the Learning Rate} Figure~\ref{fig: different learning rate} illustrates how the penalty coefficient $\lambda_t$ and the response length evolve during training under different learning rates: $\eta \in \{10^{-2}, 10^{-3}, 10^{-4}\}$.
When the learning rate is small (e.g., $\eta=10^{-4}$), we observe rapid initial reduction in response length due to the large initial penalty $\lambda_0=1e-3$. At this stage, model accuracy remains close to the reference threshold $\text{acc}_{\text{ref}}$, so $\lambda_t$ does not change significantly and continues to exert strong pressure to shorten outputs. However, as the response length is further reduced, this high penalty becomes detrimental—continued compression begins to harm accuracy. Ideally, $\lambda_t$ should decrease at this point to relax the penalty and preserve correctness. Yet with a small learning rate, the model fails to reduce $\lambda_t$ swiftly enough, and the penalty remains overly aggressive. This delayed adjustment leads to excessive shortening of outputs, ultimately causing the model to collapse in both accuracy and generation quality.

\begin{figure}[h]
    \centering
    \includegraphics[width=1\linewidth]{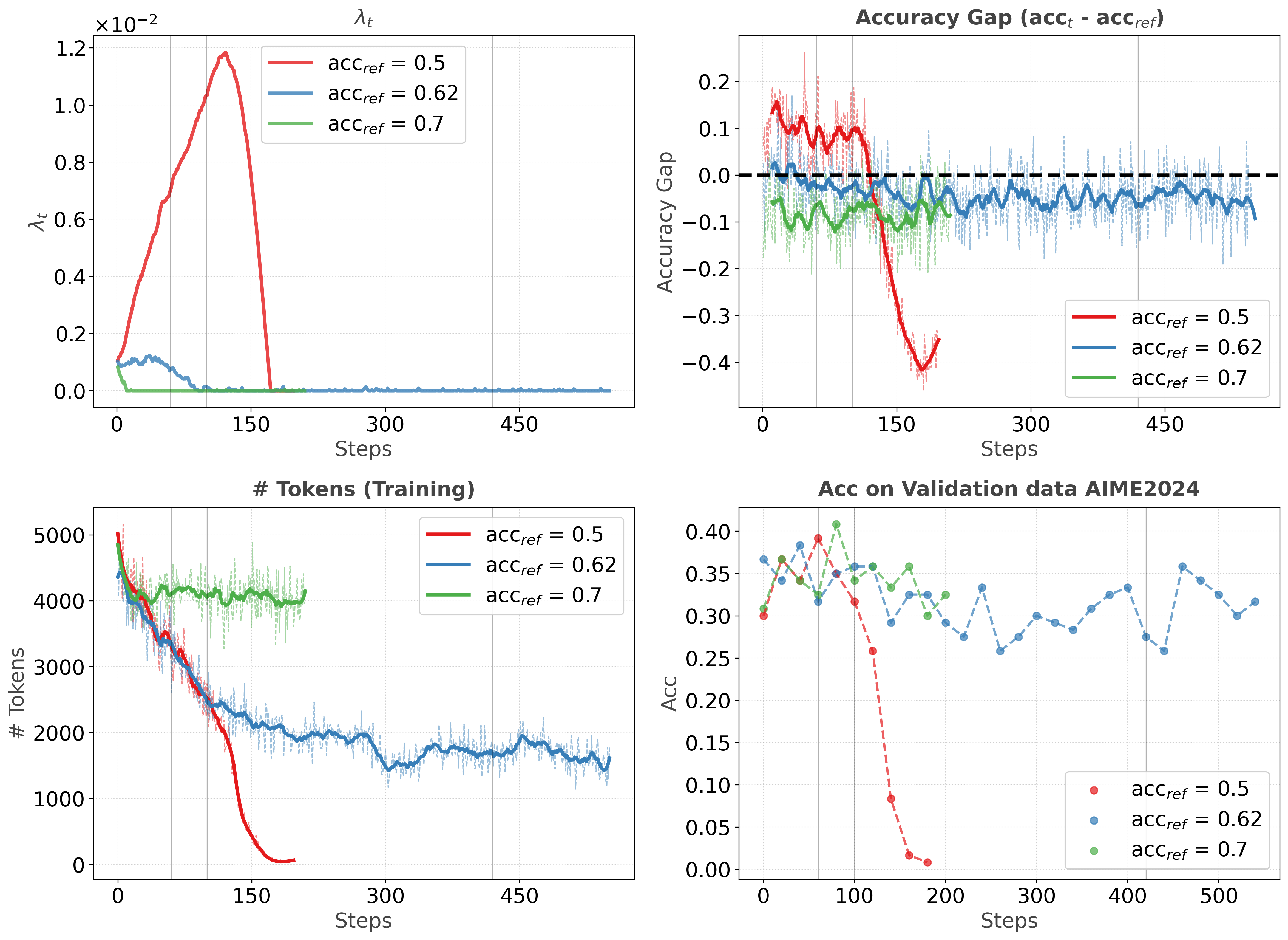}
       \caption{\textbf{Training dynamics under  different reference accuracies ($\text{acc}_{\text{ref}} \in \{0.5, 0.62, 0.7\}$).}
When the reference accuracy is set too high (e.g., 0.7), $\lambda_t$ quickly drops to zero and remains there, causing only a brief period of length reduction in the early steps and leading to convergence at a relatively high response length. When it is set too low (e.g., 0.5), $\lambda_t$ increases excessively, leading to over-penalization and eventual model collapse due to excessive shortening. }
    \label{fig: different reference acc}
\end{figure}

Conversely, when the learning rate is large (e.g., $\eta=10^{-2}$), the penalty coefficient $\lambda_t$ becomes highly sensitive to the accuracy gap, $\text{acc}_t - \text{acc}_{\text{ref}}$. Since $\text{acc}_t$ is estimated from a small batch $b_{\text{batch size}}=64$ at each step, it is inherently noisy, which amplifies fluctuations in this gap and leads to unstable updates in $\lambda_t$. For instance, starting from $\lambda_0 = 1e-3$, an accuracy drop of just 0.1 can reduce $\lambda_t$ to zero in a single step.  Consequently, $\lambda_t$ tends to oscillate sharply throughout training rather than evolving smoothly. Although the updates to $\lambda_t$ under a large learning rate are sharp and unstable—frequently spiking and then collapsing back to zero—this volatility is constrained by the non-negativity of $\lambda_t$. In particular, the penalty only activates when $\text{acc}t > \text{acc}{\text{ref}}$ and is rapidly diminished as soon as it starts harming accuracy. As a result, $\lambda_t$ often resets to zero or stays there safely, preventing prolonged over-penalization. Despite its instability, setting a relatively large learning rate can still be beneficial for length reduction: the intermittent spikes in $\lambda_t$ exert sufficient pressure to gradually shorten responses over time. At the same time, the rapid decay of $\lambda_t$ when accuracy drops acts as a built-in safeguard, preventing the model from collapsing into overly short, low-quality outputs.
\paragraph{Effect of $\text{acc}_{\text{ref}}$}
Figure~\ref{fig: different reference acc} shows the training dynamics under different reference accuracies $\text{acc}_{\text{ref}}$, selected to be either significantly higher or lower than the actual accuracy of the base model on the training data. Specifically, we use $\text{acc}_{\text{ref}}=\{0.7, 0.5\}$, which are approximately 0.1 above or below the reference accuracy used in our main experiments—estimated from the first batch of training data in a preliminary run. When the reference accuracy is set too high relative to the true accuracy of the base model (e.g., $\text{acc}_{\text{ref}} = 0.7$), the observed accuracy gap 
$\text{acc}_{t} - \text{acc}_{\text{ref}}$ remains consistently negative.  As a result, $\lambda_t$ is quickly reduced to zero within the first 10 training steps and stays there throughout. In this setting, although there is some token length reduction in the early steps (while $\lambda_t > 0$), the process plateaus quickly, and the model converges to a relatively insufficiently reduced response length. Conversely, when the reference accuracy is set too low (e.g., $\text{acc}_{\text{ref}} = 0.5$), the initial accuracy gap is consistently positive, causing $\lambda_t$ to increase steadily.
This leads to an increased gradual decline in model accuracy and a corresponding reduction in the gap $\text{acc}t - \text{acc}{\text{ref}}$, which approaches zero around step 120, where $\lambda_t$ reaches its peak. Beyond this point, the accuracy gap becomes negative, and $\lambda_t$ begins to decrease. However, similar to the case with a small learning rate, the update to $\lambda_t$ becomes too slow to respond adequately to the accuracy drop. In effect, the learning rate becomes small relative to the magnitude of $\lambda_t$, making it difficult for the penalty to relax quickly enough. As a result, the model continues to over-shorten its outputs, eventually collapsing into overly brief, low-quality responses.

\begin{figure}[h]
    \centering
    \includegraphics[width=1\linewidth]{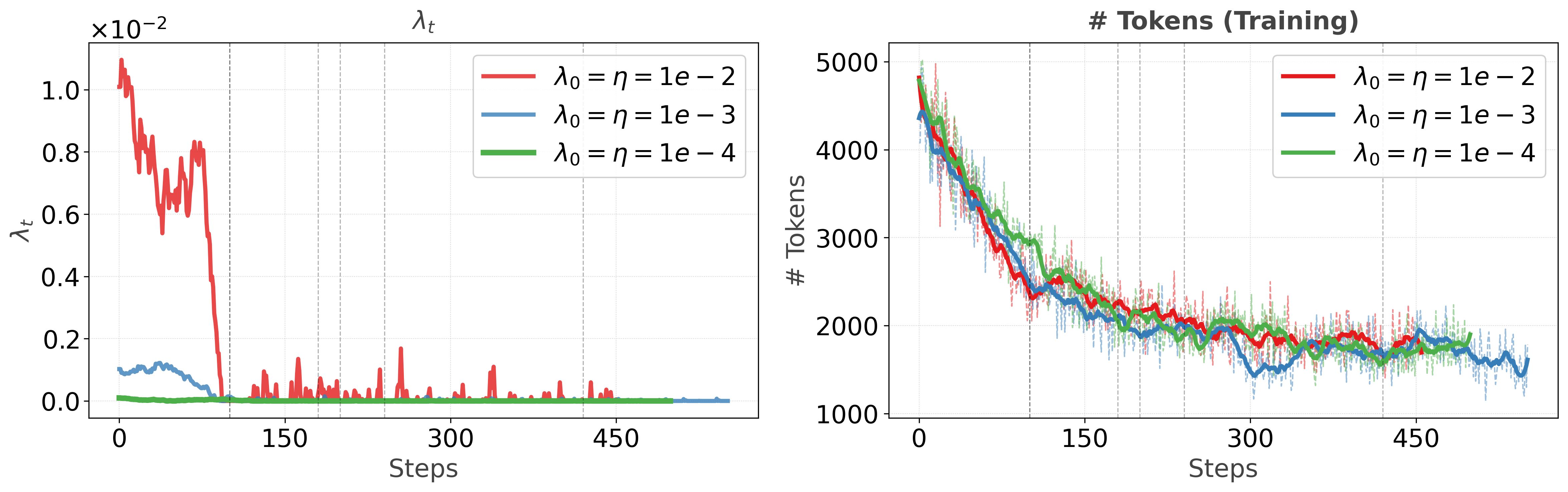}
       \caption{Training dynamics of using $\lambda_0=\eta=1e-2, 1e-3, 1e-4$ respectively.  }
    \label{fig: three sets}
\end{figure}

\paragraph{General Guidance on Parameter Setting}
From the above experiments, we can draw some general insights for practical parameter selection. When the reference accuracy estimate does not deviate significantly from the baseline model's true accuracy, the choice of learning rate should be considered in conjunction with the initial penalty strength $\lambda_0$. Specifically, we want the learning rate to be large enough to allow $\lambda_t$ to respond in a timely manner to fluctuations in accuracy—particularly when over-penalization begins to harm performance. A practical rule of thumb emerging from our experiments is as follows: the model should be able to reduce the penalty coefficient $\lambda_t$ to zero within approximately 10 steps when accuracy drops. If the maximum tolerable accuracy deviation is around 0.1, then setting the initial penalty $\lambda_0$ equal to the learning rate $\eta$ provides a reasonable balance. As for how large this shared value should be: A larger $\lambda_0$ and $\eta$ pair accelerates length reduction and reduces more aggressively, but at the cost of greater instability in $\lambda_t$ due to noisy accuracy estimates.
A smaller pair results in smoother, more stable updates to $\lambda_t$, but leads to slower and more conservative length reduction. Ultimately, the choice reflects a trade-off between reduction speed and training robustness. To illustrate this, we visualize the training dynamics under three representative settings—$(\lambda_0, \eta) \in {(10^{-2}, 10^{-2}), (10^{-3}, 10^{-3}), (10^{-4}, 10^{-4})}$—in Figure~\ref{fig: three sets}, which align with our practical guidance for parameter selection.

\section{Conclusion}
In this paper, we presented Adaptive Direct Length Penalty (A-DLP), a simple yet effective reward-shaping method that dynamically adjusts the trade-off between reasoning accuracy and response length during RL training. Unlike prior approaches that rely on static penalty terms, A-DLP adaptively adjusts the length penalty coefficient based on evolving model performance, enabling aggressive compression when safe and relaxing penalization when the length reduction becomes excessive. 
Through extensive experiments across diverse mathematical benchmarks, we show that A-DLP consistently reduces token length by over 50\% while maintaining comparable accuracy. Compared to static baselines such as S-DLP, A-DLP offers better trade-offs and avoids training collapse by stabilizing naturally as performance plateaus. Our method integrates easily into existing RL pipelines, providing a practical tool for reducing inference cost in LLMs without sacrificing quality, paving the way for building more efficient and cost-effective LLMs.
\section{Limitation and Future works}
While our method is designed to be simple, efficient, and broadly applicable, several aspects offer opportunities for future extension: first, Our experiments are conducted on a 1.5B-parameter model to ensure reproducibility and manageable compute cost. Although we expect A-DLP to generalize to larger models—especially given its lightweight integration with RL training—scaling, it remains an important next step to evaluate our methods for models of different sizes and structures. Secondly, we use a fixed learning rate for updating the penalty coefficient $\lambda_t$. While this already yields stable and interpretable behavior, a more adaptive learning rate schedule could further improve the adaptivity of the penalty to dynamic training signals and make our method more robust across diverse settings. Thirdly, A-DLP reduces response lengths in a roughly proportional manner across correct and incorrect responses. While this leads to overall efficiency gains, in the future, we plan to improve our method to compress incorrect outputs more aggressively to better preserve the accuracy of correct ones.
\bibliography{neurips}
\bibliographystyle{plain}

\newpage
\appendix
\section{Additional Experiments}
\begin{figure}[h]
    \centering
    \includegraphics[width=1\linewidth]{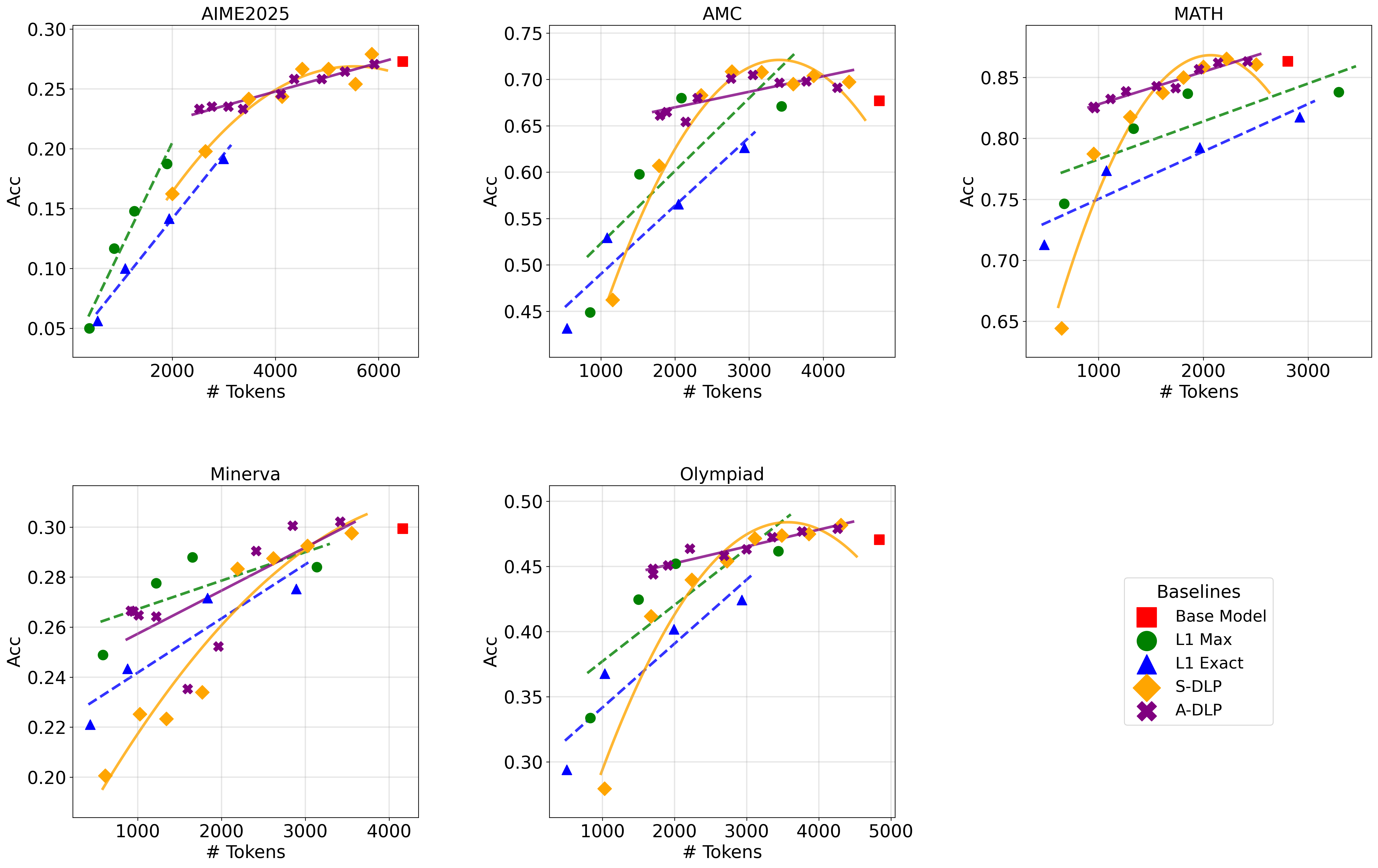}
    \caption{\textbf{Performance comparison of A-DLP with baselines using more configurations.} For A-LDP, we plot the evaluation results for different checkpoints during training. For L1-Max and L1-Exact, we plot different target token budgets. }
    \label{fig: main-more configurations}
\end{figure}

\begin{figure}[h]
    \centering
    \includegraphics[width=1\linewidth]{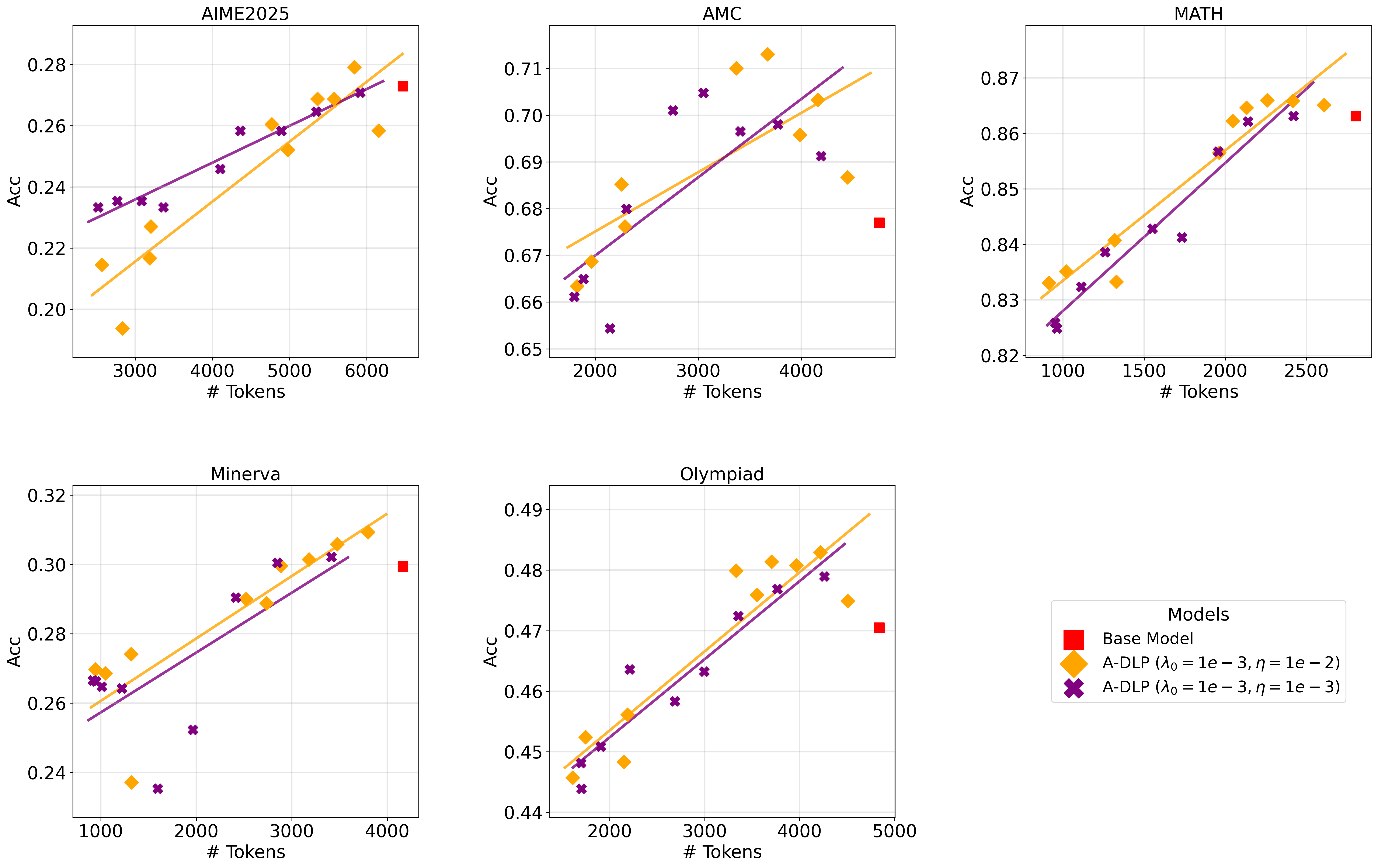}
    \caption{\textbf{Accuracy and token length trade-off comparison of A-DLP on two sets of parameters.} }
    \label{fig: different params}
\end{figure}

\begin{figure}[h]
    \centering
    \includegraphics[width=1\linewidth]{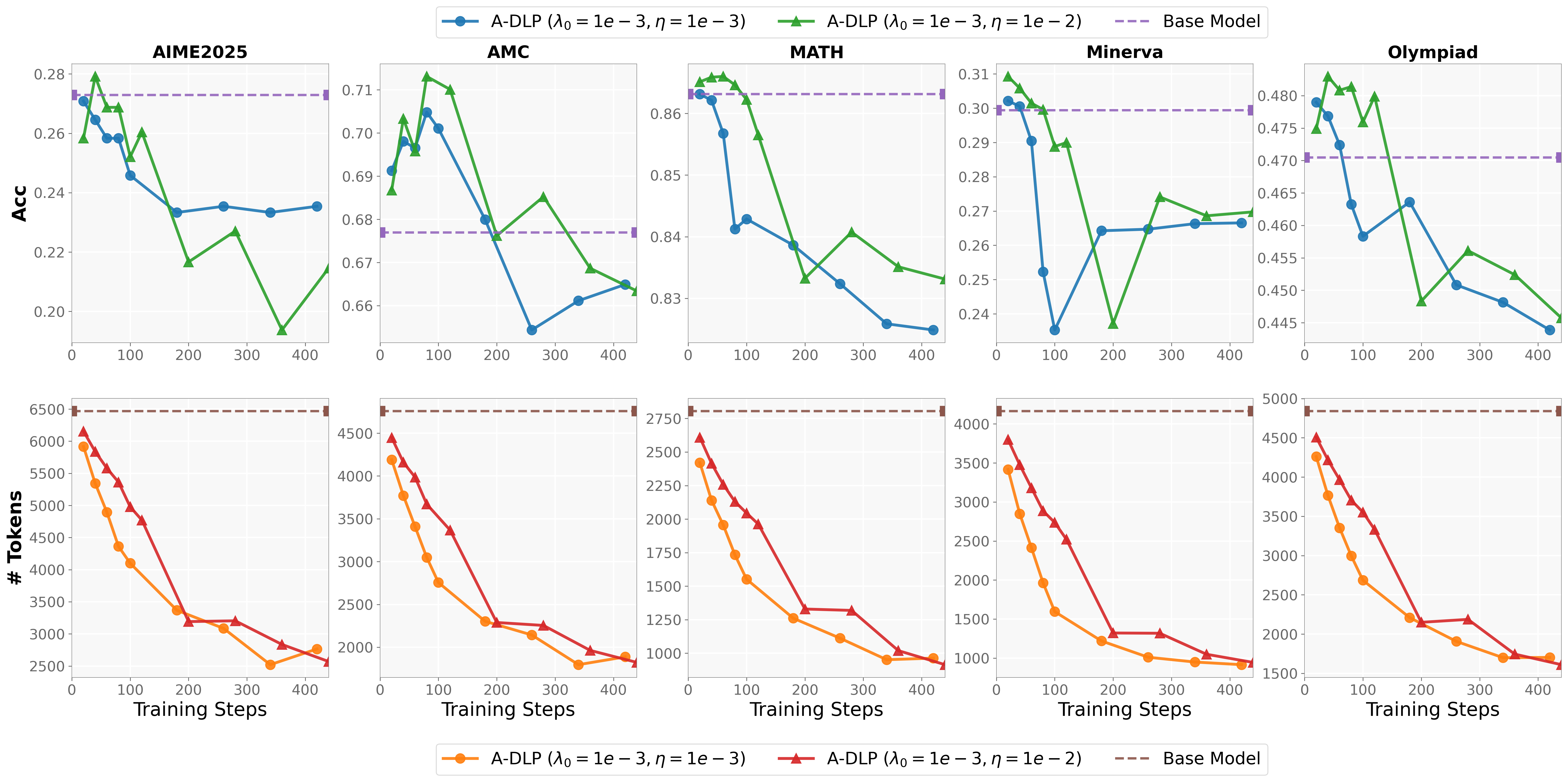}
    \caption{\textbf{Accuracy and token length comparison of A-DLP on two sets of parameters varying on the training steps.} }
    \label{fig: different params-step wise}
\end{figure}

\begin{figure}[h]
    \centering
    \includegraphics[width=1\linewidth]{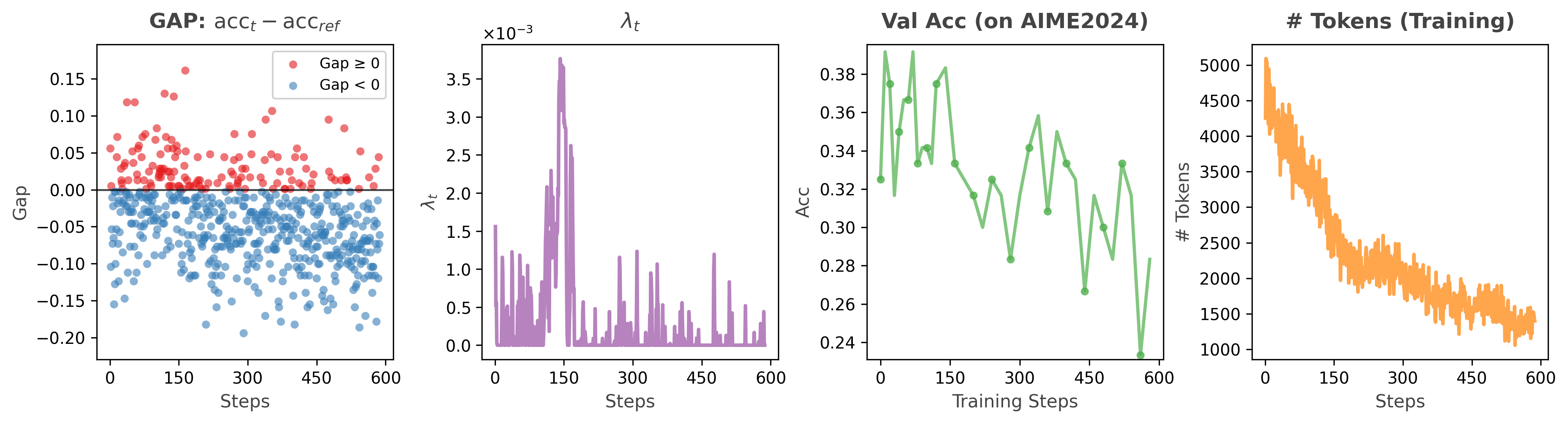}
    \caption{\textbf{Training behavior using a larger learning rate $\eta=1e-2$.} }
    \label{fig: training dynamics 1e-2}
\end{figure}

In Figure \ref{fig: main-more configurations}, we plot the A-DLP, L1-Max and L1-Exact with more configurations and fit a line over these configurations. For both L1-Max and L1-Exact, we use the budget to be $\{512, 1024, 2048, 4096\}$ while the max token length set to be twice of the budget. While, for A-DLP, we plot the evaluation result for different checkpoints during training. 
In Figure \ref{fig: different params}, we compare A-DLP on its accuracy and token length trade-off using two sets of parameters, $\lambda_0=1e-3, \eta=1e-2$ and $\lambda_0=1e-3, \eta=1e-3$ on the testing data over different training steps. 
Figure \ref{fig: different params-step wise} compares accuracy and token length respectively with x-axis being training steps. Figure \ref{fig: training dynamics 1e-2} shows the training behavior of using parameter set $\lambda_0=1e-3, \eta=1e-2$. 

\begin{figure}[h]
    \centering
    \includegraphics[width=1\linewidth]{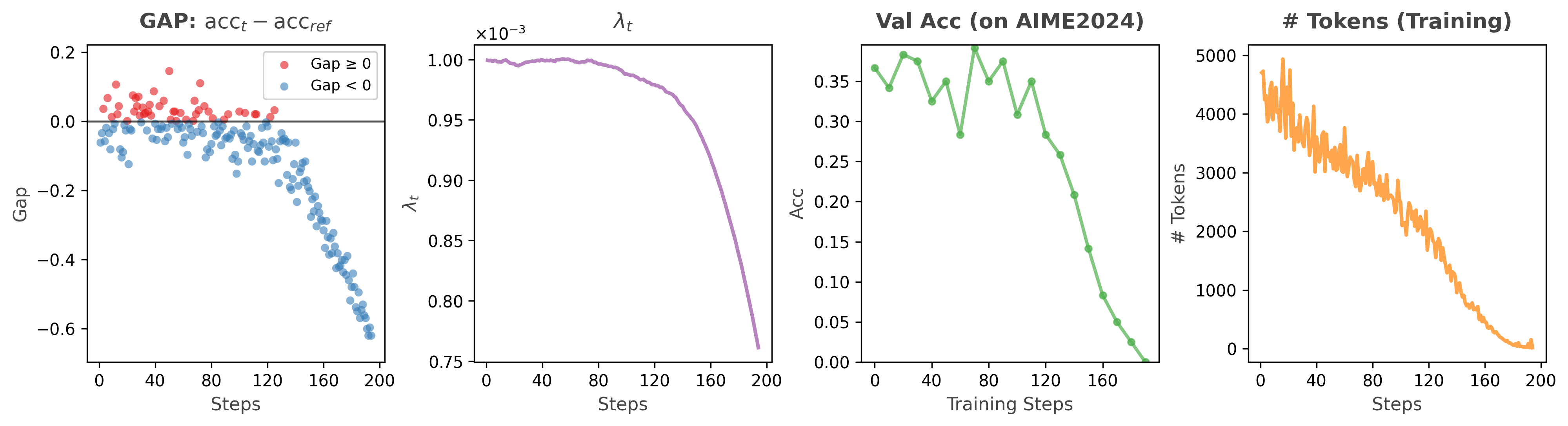}
    \caption{\textbf{Training behavior using a smaller learning rate $\eta=1e-4$.} }
    \label{fig: training dynamics 1e-4}
\end{figure}

\begin{figure}[h]
    \centering
    \includegraphics[width=1\linewidth]{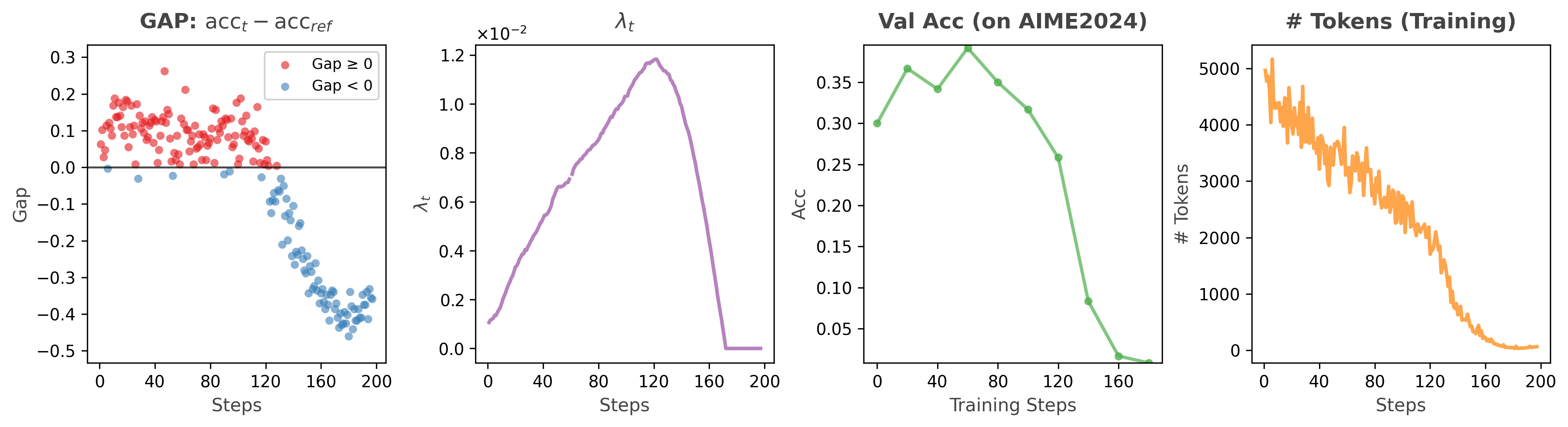}
    \caption{\textbf{Training behavior using a reference accuracy $\text{acc}_{\text{ref}}=0.5$ that is much smaller than the accuracy of the base model.} }
    \label{fig: training dynamics acc 0.5}
\end{figure}

\begin{figure}[h]
    \centering
    \includegraphics[width=1\linewidth]{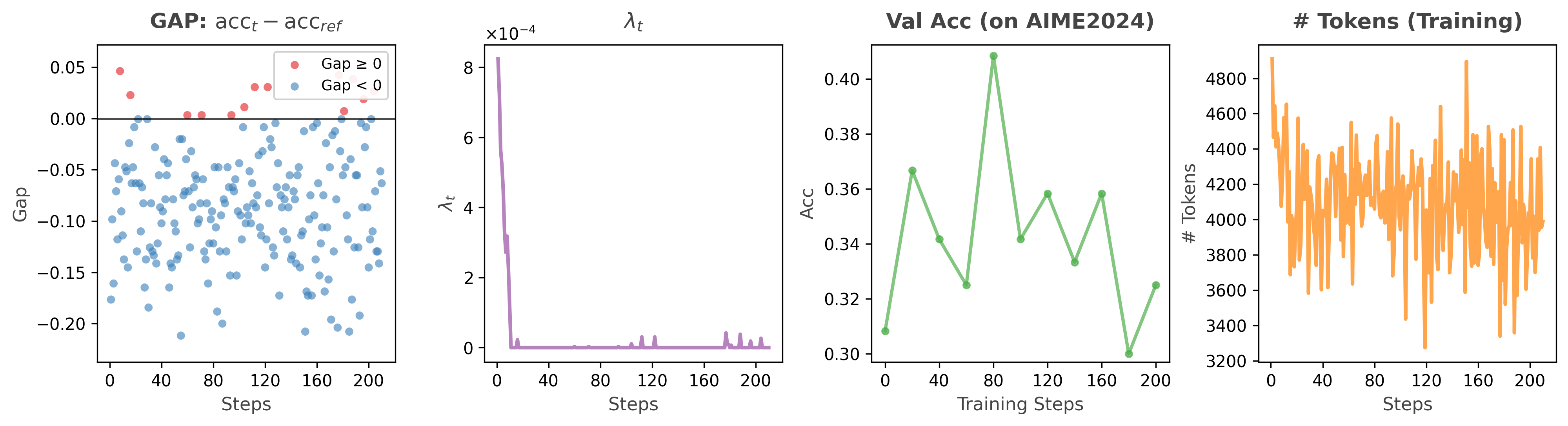}
    \caption{\textbf{Training behavior using a reference accuracy $\text{acc}_{\text{ref}}=0.7$ that is much larger than the accuracy of the base model.} }
    \label{fig: training dynamics acc 0.7}
\end{figure}
\section{Effect of $\lambda_0$}
\begin{figure}[h]
    \centering
    \includegraphics[width=1\linewidth]{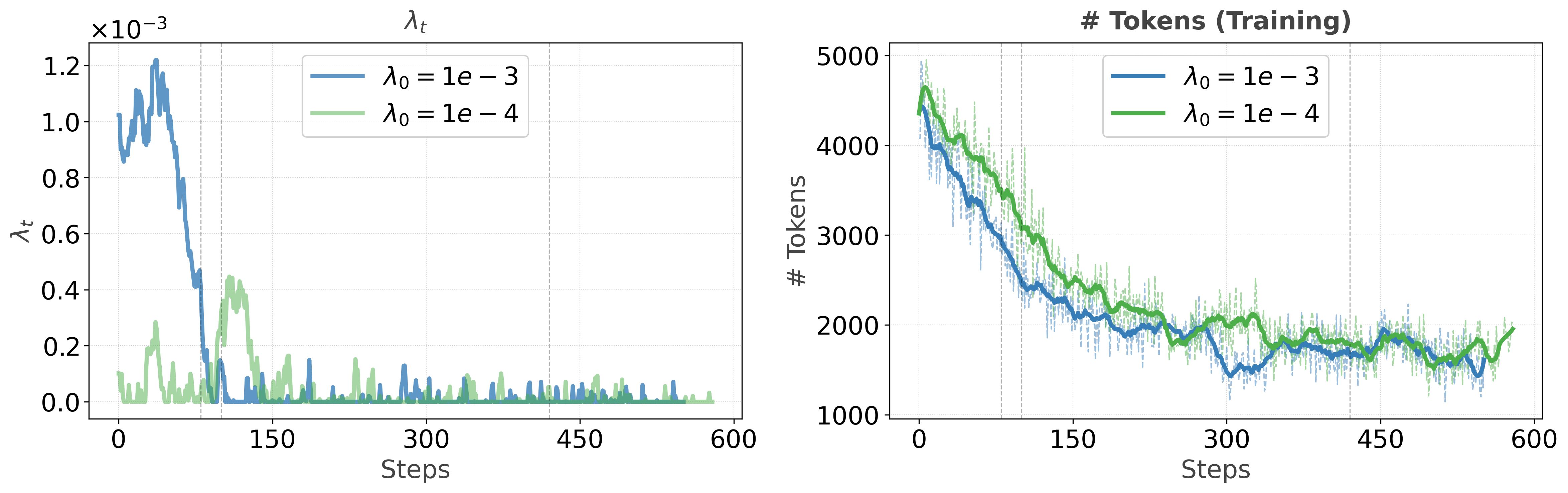}
    \caption{Training dynamics of using different $\lambda_0$ while keeping learning rate $\eta=1e-3$. }
    \label{fig: different lambda0}
\end{figure}

In Figure \ref{fig: different lambda0}, we show the effect of different initialization $\lambda_0=\{1e-3, 1e-4\}$. 

\end{document}